%% file: cement.tex
\def\year{2020}\relax
\documentclass[letterpaper]{article} 
\usepackage{aaai20}  
\usepackage{times}  
\usepackage{helvet}  
\usepackage{courier}  
\usepackage{array}
\usepackage{mathtools}
\usepackage{multirow}
\usepackage{enumitem}
\usepackage{dsfont}
\usepackage{url}  
\usepackage{graphicx, subfigure}
\usepackage{amsmath}
\usepackage{amsfonts}
\usepackage{bm}
\usepackage{amssymb}  
\usepackage[draft]{todonotes}
\newcommand{\nosection}[1]{\vspace{2pt}\noindent\textbf{#1.}}

\usepackage[linesnumbered,boxed,ruled]{algorithm2e}
\begin{document}
\title{Industrial Scale Privacy Preserving Deep Neural Network}

\author{
Longfei Zheng, Chaochao Chen$\thanks{Corresponding author}$, Yingting Liu, Bingzhe Wu, Xibin Wu, Li Wang, Lei Wang, \\
{\Large{\textbf{Jun Zhou}, \textbf{Shuang Yang}}} \\
Ant Financial Services Group\\
\{zlf206411,chaochao.ccc,yingting,fengyuan.wbz,xibin.wxb,raymond.wangl,shensi.wl,jun.zhoujun,shuang.yang\}@antfin.com
}

\maketitle

\begin{abstract}
Deep Neural Network (DNN) has been showing great potential in kinds of real-world applications such as fraud detection and distress prediction. 
Meanwhile, data isolation has become a serious problem currently, i.e., different parties cannot share data with each other. 
To solve this issue, most research leverages cryptographic techniques to train secure DNN models for multi-parties without compromising their private data. 
Although such methods have strong security guarantee, they are difficult to scale to deep networks and large datasets due to its high communication and computation complexities. 
To solve the scalability of the existing secure Deep Neural Network (DNN) in data isolation scenarios, in this paper, we propose an industrial scale privacy preserving neural network learning paradigm, which is secure against semi-honest adversaries. 
Our main idea is to split the computation graph of DNN into two parts, i.e., the computations related to private data are performed by each party using cryptographic techniques, and the rest computations are done by a neutral server with high computation ability. 
We also present a defender mechanism for further privacy protection. 
We conduct experiments on real-world fraud detection dataset and financial distress prediction dataset, the encouraging results demonstrate the practicalness of our proposal. 
\end{abstract}
\input{section/intro}
\input{section/relatedwork}
\input{section/model}
\input{section/experiment}

\input{section/conclusion}

\bibliographystyle{aaai}
\bibliography{cement}

\end{document}

%% file: section/intro.tex
\section{Introduction}	

\begin{figure}[t]
\centering
\includegraphics[width=3.6cm]{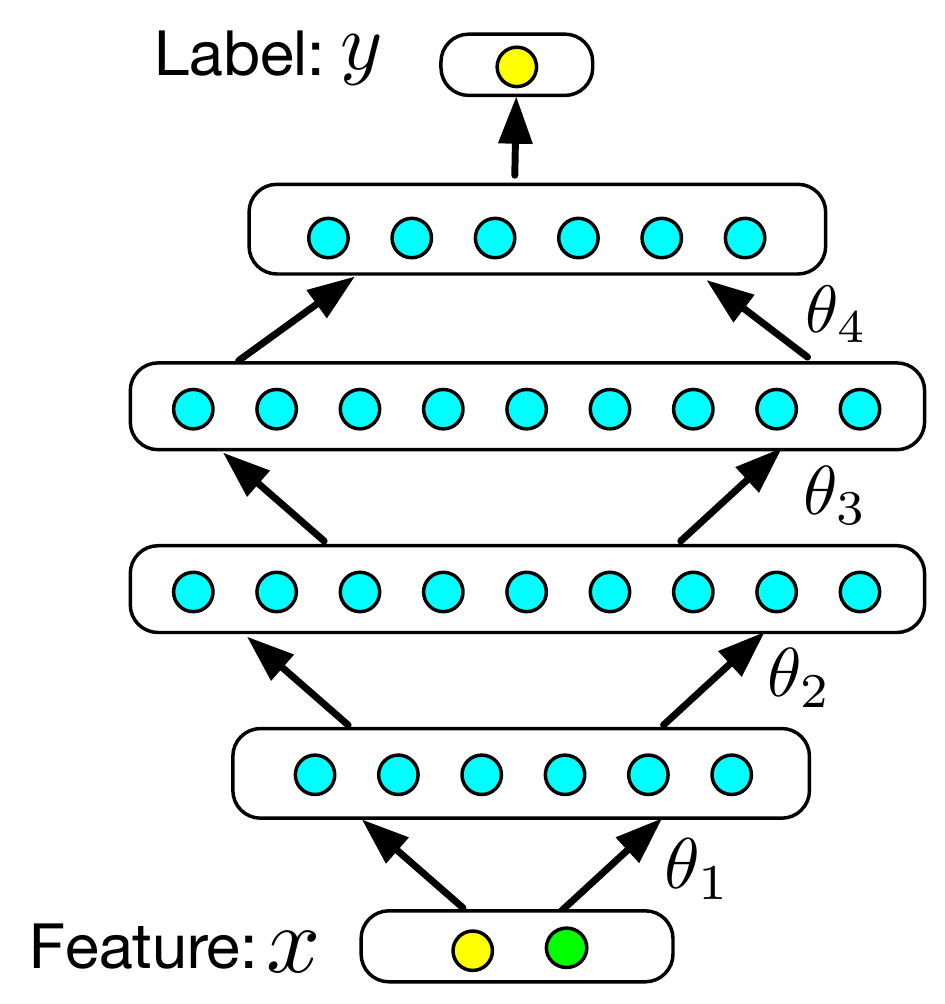}
\caption{Structure of a traditional neural network.}
\label{nnexample}
\end{figure}

Deep Neural Network (DNN) has been showing great potential in kinds of machine learning tasks and successfully applying in various applications such as computer vision \cite{howard2017mobilenets},  recommender system \cite{zhu2019dtcdr}, and financial distress prediction \cite{chen2008NN}, due to its powerful representation ability. 
Deep learning allows computational models that are composed of multiple processing layers to learn representations of data with multiple levels of abstraction \cite{lecun2015deep}. 
The general structure of a neural network is shown in Figure \ref{nnexample}. 
Meanwhile, data isolation has become a serious problem currently, especially with kinds of national data protection regulations coming into force. 
That is, different organizations (parties) are reluctant or cannot share sensitive data with each other due to competition or regulation reasons. 
Such data isolation problem has limited the power of DNN, since DNN usually achieves better performance with more data. 

To solve this problem, existing researches adopted cryptographic techniques, e.g., homomorphic encryption \cite{gilad2016cryptonets} or secure
multi-party computation \cite{mohassel2017secureml}, for multi-parties to train privacy preserving neural networks. 
Although such cryptographic based neural networks have strong security guarantee, they are difficult to scale to deep network structures and large datasets due to its high communication and computation complexities. 
However, in industry, the real-world applications happened to have two characteristics: 
(1) the datasets are large due to millions of data are hold by big companies, and 
(2) the neural network structures are deep so as to learn the patterns in big data. 
Therefore, efficiency becomes a main challenge when applying existing privacy preserving neural networks in practice. 

To ingeniously balance the privacy and scalability, in this paper, we propose an industrial-scale privacy preserving neural network learning paradigm, which is one of the \textit{shared machine learning algorithms} in Ant Financial \cite{chen2018privacy,chen2020secure,chen2020practical,liu2020privacy}. 
Motivated by split learning \cite{vepakomma2018split}, we split the computation graph of DNN into two kinds, i.e., the computations related to private data are performed by each party using cryptographic techniques, and the rest computations are done by a neutral server with high computation ability. 
To this end, both private data and model are hold by data holders, and the heavy non-private data related computations are done by a neutral server. 
To further protect data privacy, we propose a defender mechanism when training the model so that the neural server cannot infer the private input of data holders from the hidden layers. 
Therefore, our proposal not only preserves data private, but also has good scalability. 
We conduct experiments on real-world fraud detection and distress prediction datasets, the results demonstrate that our proposed privacy preserving neural network has almost the same performance with the traditional neural model. 

Our main contributions are summarized as follows:
\begin{itemize} [leftmargin=*] \setlength{\itemsep}{-\itemsep}
\item We propose an industrial-scale privacy preserving neural network learning paradigm with defender mechanism, which not only preserves data privacy, but also has good scalability. 
\item We implement our model on decentralized network settings, where computation nodes have their own private data and they can train privacy preserving neural network models.
\item Our proposal is verified on real-world datasets and the results show its superiority.
\end{itemize}

%% file: section/relatedwork.tex
\section{Related Work}\label{background}
We first simply review deep learning models and then describe two popular types of privacy preserving neural network models. 

\nosection{Deep learning}
Deep Neural Network (DNN) has been showing great power in kinds of machine learning tasks, since it can learn complex functions by composing multiple non-linear modules to transform representations from low-level raw inputs to high-level abstractions \cite{gu2019securing}. 
Mathematically, the forward procedure of a DNN can be defined as a representation function $f$ that maps an input $\textbf{X}$ to an output $y$, i.e., $y=f(\textbf{X}, \bm\theta)$, where $\bm\theta$ is model parameter. 
Assume a DNN has $L$ layers, then $f$ is composed of $L$ sub-functions $f_{l|{l\in[1,L]}}$, which are connected in a chain. That is, $f(\textbf{X})=f_Lf_{L-1}...f_1(\textbf{X}, \bm\theta_1)$, as is shown in Figure \ref{nnexample}.

\nosection{Cryptographic based methods}
These methods use cryptographic techniques, e.g., secret sharing and homomorphic encryption, to build approximated neural networks models \cite{mohassel2017secureml,wagh2019securenn}, since the nonlinear active functions are not cryptographically computable. 
These models are difficult to scale to deep networks and large datasets due to the high communication and computation complexities of the cryptographic techniques. 
In this paper, we use cryptographic techniques for data holders to calculate the hidden layers securely.

\nosection{Split neural graph based methods} 
These methods split the computation graph of neural networks into two parts, i.e., let data holders calculate the private data related computations individually and get a hidden layer, and then let a server makes the rest computations \cite{gupta2018distributed,vepakomma2018split,osia2019hybrid,gu2019securing}. 
For example, Gu et al. \cite{gu2019securing} proposed to enclose sensitive computation in a trusted execution environment, i.e., Intel Software Guard Extensions \cite{mckeen2013innovative}, to mitigate input information disclosures, and then delegate non-sensitive workloads with hardware-assisted deep learning acceleration. 
Our model differs from them in mainly two aspects. 
First, we use cryptographic techniques for data holders to calculate the hidden layers collaboratively rather than compute them based on their plaintext data individually. 
Seconds, we propose a defender mechanism when training the model so that the neutral server cannot infer the private input of data holders from the hidden layers. Therefore, our model has better privacy guarantee. 

%% file: section/model.tex
\section{The Proposed Method}

\subsection{Problem Description}
We start from a concrete example. 
Suppose there are two financial companies, i.e., $\mathcal{A}$ and $\mathcal{B}$, who both need to detect fraud users. 
$\mathcal{A}$ has some user features ($\textbf{X}_A$) and labels ($\textbf{y}_A$), and $\mathcal{B}$ has features ($\textbf{X}_B$) for the same batch of users. 
Although $\mathcal{A}$ can build a Deep Neural Network (DNN) for fraud detection using its own data, the model performance can be improved by incorporating features of $\mathcal{B}$. 
However, these two companies can not share data with each other due to the fact that leaking users' private data is against regulations. 
This is a classic data isolation problem. 
It is challenging for both parties to build a privacy preserving neural network collaboratively without compromising their private data. 
In this paper, we only consider the situation where two data holders have the same sample set, one of them ($\mathcal{A}$) has partial features and labels, and the other ($\mathcal{B}$) has the rest partial features. 
Our proposal can be naturally extended to more than two parties. 

\begin{figure}[t]
\centering
\includegraphics[width=6cm]{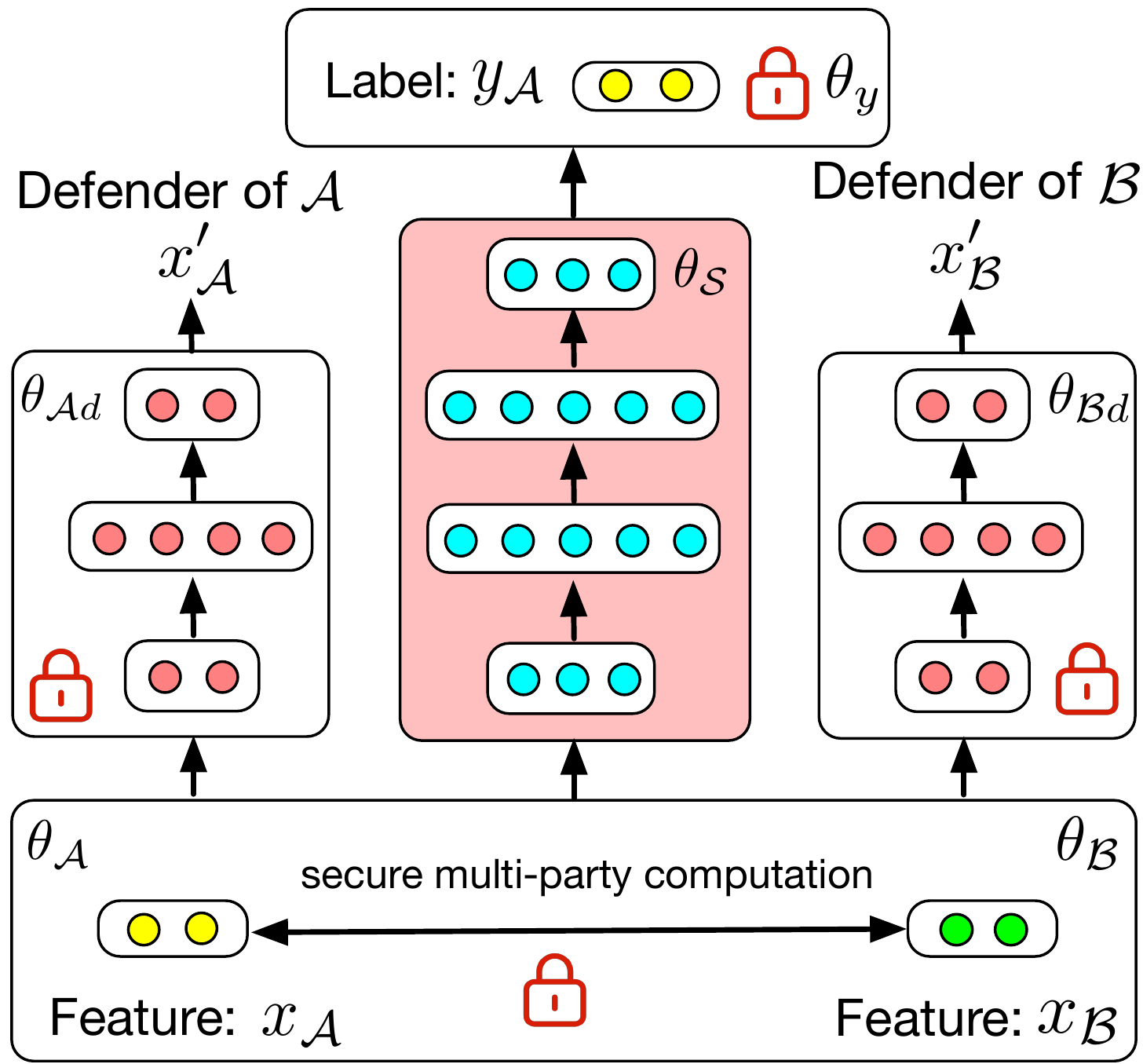}
\caption{The proposed privacy preserving neural network. The middle pink part is performed on a neutral server and the rest are done by data holders.}
\label{framework}
\end{figure}

\subsection{Proposal Overview}
We propose a novel privacy preserving neural network learning framework for the above challenge. 
As described in related work, DNN can be defined as a layer-wise representation function. 
Motivated by the existing work \cite{gupta2018distributed,vepakomma2018split,osia2019hybrid,gu2019securing}, we propose to decouple the computation graph of DNN into two kinds, i.e., the computations related to private data are performed by each party using cryptographic techniques, and the rest computations are done by a neutral server with high computation ability. 
Here, the private data are the input and output of the neural network, which corresponding to the private features and labels from data holders. 

Specifically, we divide the model parameters ($\theta$) into three parts, \textit{the computations that are related to private features on both data holders} ($\theta_\mathcal{A}$ and $\theta_\mathcal{B}$), \textit{the computations related to private labels on a data holder} ($\theta_y$), and \textit{the rest heavy hidden layer related computations on a neutral server} ($\theta_\mathcal{S}$). 
As shown in Figure \ref{framework}, the first two parts are private data related computations and therefore are performed by data holders themselves using secure multi-party computation techniques, and the rest computations can be done by a neutral server which has rich computation resources. 
Moreover, as has been pointed out by literature, attackers may recover the raw input data given the hidden layers of a DNN. To prevent the neutral server inferring the private input of data holders from the hidden layers, we propose a \textit{defender mechanism} when training the model. 
Our solution is against \textit{semi-honest adversary}, i.e., the corrupted participants will still strictly follow the protocol but may want to learn more information. 
We will describe each module in details in the following subsections.

\begin{algorithm*}[h]
\label{securelr}
\caption{Data holders $\mathcal{A}$ and $\mathcal{B}$ securely calculate the first hidden layer using secret sharing}\label{algo-horLR}
\KwIn {features of $\mathcal{A}$ and $\mathcal{B}$ ($\textbf{X}_A$ and $\textbf{X}_B)$ and current models of $\mathcal{A}$ and $\mathcal{B}$ ($\bm{\theta}_A$ and $\bm{\theta}_B)$}
\KwOut{The share of first hidden layer for $\mathcal{A}$ and $\mathcal{B}$}

$\mathcal{A}$ and $\mathcal{B}$ locally generate $\left\langle\textbf{X}_A\right\rangle_1$ and $\left\langle\textbf{X}_A\right\rangle_2$, and $\left\langle\textbf{X}_B\right\rangle_1$ and $\left\langle\textbf{X}_B\right\rangle_2$, respectively\\
$\mathcal{A}$ and $\mathcal{B}$ locally generate $\left\langle\bm{\theta}_A\right\rangle_1$ and $\left\langle\bm{\theta}_A\right\rangle_2$, and $\left\langle\bm{\theta}_B\right\rangle_1$ and $\left\langle\bm{\theta}_B\right\rangle_2$, respectively\\
$\mathcal{A}$ distributes $\left\langle\textbf{X}_A\right\rangle_2$ and $\left\langle\bm{\theta}_A\right\rangle_2$ to $\mathcal{B}$ \\
$\mathcal{B}$ distributes $\left\langle\textbf{X}_B\right\rangle_1$ and $\left\langle\bm{\theta}_B\right\rangle_1$ to $\mathcal{A}$ \\
$\mathcal{A}$ locally calculates $\left\langle\textbf{X}\right\rangle_1 = \left\langle\textbf{X}_A\right\rangle_1 \oplus \left\langle\textbf{X}_B\right\rangle_1$, $\left\langle\bm{\theta}\right\rangle_1 = \left\langle\bm{\theta}_A\right\rangle_1 \oplus \left\langle\bm{\theta}_B\right\rangle_1$, and $\left\langle\textbf{X}\right\rangle_1 \times \left\langle\bm{\theta}\right\rangle_1$ \\
$\mathcal{B}$ locally calculates $\left\langle\textbf{X}\right\rangle_2 = \left\langle\textbf{X}_A\right\rangle_2 \oplus \left\langle\textbf{X}_B\right\rangle_2$, $\left\langle\bm{\theta}\right\rangle_2 = \left\langle\bm{\theta}_A\right\rangle_2 \oplus \left\langle\bm{\theta}_B\right\rangle_2$, and $\left\langle\textbf{X}\right\rangle_2 \times \left\langle\bm{\theta}\right\rangle_2$ \\

$\mathcal{A}$ and $\mathcal{B}$ calculate $\left\langle\textbf{X}\right\rangle_1 \times \left\langle\bm{\theta}\right\rangle_2$ and $\left\langle\textbf{X}\right\rangle_2 \times \left\langle\bm{\theta}\right\rangle_1$ using secret sharing matrix multiplication, $\mathcal{A}$ get $\left\langle\textbf{X}_1 \times \bm{\theta}_2\right\rangle_A $ and $\left\langle\textbf{X}_2 \times \bm{\theta}_1\right\rangle_A $, $\mathcal{B}$ gets $\left\langle\textbf{X}_1 \times \bm{\theta}_2\right\rangle_B $ and $\left\langle\textbf{X}_2 \times \bm{\theta}_1\right\rangle_B $\\ 

$\mathcal{A}$ locally calculates $\left\langle\textbf{X} \times \bm{\theta}\right\rangle_A = \left\langle\textbf{X}\right\rangle_1 \times \left\langle\bm{\theta}\right\rangle_1 + \left\langle\textbf{X}_1 \times \bm{\theta}_2\right\rangle_A  + \left\langle\textbf{X}_2 \times \bm{\theta}_1\right\rangle_A$ \\

$\mathcal{B}$ locally calculates $\left\langle\textbf{X} \times \bm{\theta}\right\rangle_B = \left\langle\textbf{X}\right\rangle_2 \times \left\langle\bm{\theta}\right\rangle_2 + \left\langle\textbf{X}_1 \times \bm{\theta}_2\right\rangle_B  + \left\langle\textbf{X}_2 \times \bm{\theta}_1\right\rangle_B$ \\

\Return $\left\langle\textbf{X} \times \bm{\theta}\right\rangle_A$ for $\mathcal{A}$ and $\left\langle\textbf{X} \times \bm{\theta}\right\rangle_B$ for $\mathcal{B}$
\end{algorithm*}

\subsection{Private Feature Related Computations}

Private feature related computations refer to data holders collaboratively calculate the hidden layer of a DNN using their own private data. 
Here, data holders want to (1) calculate a common function, i.e., $f_1(\textbf{X}_A, \textbf{X}_B)$, collaboratively and (2) keep their features, i.e., $\textbf{X}_A$ and $\textbf{X}_B$, private. 
Secure multi-party computation \cite{yao1982protocols} was born to solve this problem. 
Mathematically, $\mathcal{A}$ and $\mathcal{B}$ have partial features ($\textbf{X}_A$ and $\textbf{X}_B$) and partial model parameters ($\bm{\theta}_A$ and $\bm{\theta}_B$), respectively, and they want to compute the output of the first hidden layer collaboratively. 
That is, $\mathcal{A}$ and $\mathcal{B}$ want to compute 
$\textbf{h}_1 = f_1(\textbf{X}_A \oplus \textbf{X}_B, \bm{\theta}_A \oplus \bm{\theta}_B)$, 
where $\oplus$ denotes concatenation operation and $f_1$ is the active function.

We propose to solve the above problem using secret sharing \cite{shamir1979share}. 
The main technique used is secret sharing based matrix addition and multiplication on fixed-point numbers. 
Please refer to \cite{mohassel2017secureml} for more details. 
Assuming $f_1$ is a linear active function, we propose a secure protocol in Algorithm 1. 
Note that the nonlinear active functions can be approximated by using polynomials or Taylor expansion \cite{hardy2017private}. 
To this end, $\mathcal{A}$ and $\mathcal{B}$ each obtains a partial share of the hidden layer, i.e., $\textbf{h}_1 = \langle \textbf{h}_1 \rangle _A + \langle \textbf{h}_1 \rangle _B$.

\subsection{Hidden Layer Related Computations}
After $\mathcal{A}$ and $\mathcal{B}$ obtain the shares of the first hidden layer, they send them to a neutral server for hidden layer related computations, i.e., $\textbf{h}_{L} = f (\textbf{h}_l, \bm{\theta}_{S})$. 
This is the same as the traditional neural networks. 
Given $l$-th hidden layer $\textbf{h}_l$, where $1 \le l \le L-1$ and $L$ be the number of hidden layers, the $(l+1)$-th hidden layer can be calculated by 
\begin{equation}\label{hdl}
\textbf{h}_{l+1} = f_l (\textbf{h}_l, \bm{\theta}_{l}),
\end{equation}
where $\bm{\theta}_{l}$ is the parameters in $l$-th layer, and $f_l$ is the active function of the $l$-th layer. 
These are the most time-consuming computations, because there are many nonlinear operations, e.g., max pooling, are not cryptographically
friendly. We leave these heavy computations on a neutral server who has strong computation power. 
To this end, our model can scale to large dataset. 

\subsection{Private Label Related Computations}
After the neutral server finishes the hidden layer related computations, it sends the final hidden layer $\textbf{h}_L$ to the data holder who has the label, i.e., $\mathcal{A}$ in this case, for computing predictions. That is 
\begin{equation}\label{pre}
\hat{\textbf{y}} = \delta (\textbf{h}_L, \bm{\theta}_{L}),
\end{equation}
where $\delta$ is designed based on different prediction tasks, e.g., $\delta$ be the logistic function for a binary classification task. 

\begin{algorithm*}[t]
\label{algo}
\caption{Privacy preserving neural network}
\KwIn {Features of $\mathcal{A}$ ($\textbf{X}_A$), features of $\mathcal{B}$ ($\textbf{X}_B$), a neutral server ($\mathcal{S}$), and the number of iteration ($T$)}
\KwOut{Trained model ($\bm{\theta}$) and defender ($\bm{\theta}_{\mathcal{A}d}$ and $\bm{\theta}_{\mathcal{B}d}$)}

$\mathcal{A}$, $\mathcal{B}$, and the neutral server initialize model parameters \\

\For{$t=1$ to $T$}
{
	\For{each mini-batch in training datasets}
	{
		\# Forward computation \\
		$\mathcal{A}$ and $\mathcal{B}$ collaboratively learn the first hidden layer based on Algorithm 1 and send the result to $\mathcal{S}$\\
		$\mathcal{A}$ and $\mathcal{B}$ calculate the recovered input by $f(\textbf{h}_1, \bm{\theta}_{\mathcal{A}d})$ and $f(\textbf{h}_1, \bm{\theta}_{\mathcal{B}d})$, respectively \\
		$\mathcal{S}$ calculates the rest hidden layers by $\textbf{h}_{L} = f (\textbf{h}_l, \bm{\theta}_{S})$ \\
		$\mathcal{S}$ sends $\textbf{h}_L$ back to $\mathcal{A}$ \\
		$\mathcal{A}$ makes predictions by Eq. \eqref{pre} \\
		\# Backward computation \\
		Update model parameters $\bm{\theta}$, including $\bm{\theta}_A$, $\bm{\theta}_B$, $\bm{\theta}_S$, and $\bm{\theta}_y$, using the gradient 
		$\Delta_{\bm{\theta}}$Eq. \eqref{obj} \\
		Update defender parameters $\bm{\theta}_d$ using the gradient $\Delta_{\bm{\theta}_{\mathcal{A}d}}$Eq. \eqref{defender1} and $\Delta_{\bm{\theta}_{\mathcal{B}d}}$Eq. \eqref{defender2}
	}
}
\Return Trained model ($\bm{\theta}$) and defender ($\bm{\theta}_{\mathcal{A}d}$ and $\bm{\theta}_{\mathcal{B}d}$)
\end{algorithm*}

\subsection{Strengthening Privacy with Defender Mechanism}
To further protect data privacy, we propose a defender mechanism when training the model so that the neural server cannot infer the private input of data holders from the hidden layers. 
As can be seen in Figure \ref{framework}, the defender tries to learn a representation that maps hidden layer to the private input (features), just as an attacker would do. 
Given the hidden layer ($\textbf{h}_1$), the recovered input of $\mathcal{A}$ and $\mathcal{B}$ are $f(\textbf{h}_1, \bm{\theta}_{\mathcal{A}d})$ and $f(\textbf{h}_1, \bm{\theta}_{\mathcal{B}d})$, respectively. 
Therefore, to protect the private features being recovered, the defender losses of $\mathcal{A}$ and $\mathcal{B}$ are 
\begin{equation}\label{defender1}
\mathop {\max }\limits_{\bm{\theta}_{\mathcal{A}d}} d(\textbf{X}_A, f(\textbf{h}_1, \bm{\theta}_{\mathcal{A}d})),
\end{equation}
\begin{equation}\label{defender2}
\mathop {\max }\limits_{\bm{\theta}_{\mathcal{B}d}} d(\textbf{X}_B, f(\textbf{h}_1, \bm{\theta}_{\mathcal{B}d})),
\end{equation}
where $\bm{\theta}_{\mathcal{A}d}$ and $\bm{\theta}_{\mathcal{B}d}$ are the defender model of $\mathcal{A}$ and $\mathcal{B}$ and $d(\cdot,\cdot)$ measures the distance between original input and recovered input. 
With the present of the defender, it becomes difficult for the server to infer the input given the hidden layer and the corresponding input. 
We will empirically study the effect of the defender on privacy protection in experiments. 

\subsection{Putting All together}
Our model consists of the private feature related computations, hidden layer related computations, private label related computations, and the defender. 
The first three parts compute the output loss of the DNN, and the last part is the defender loss.  
Thus, the total loss becomes
\begin{equation}\label{obj}
\mathcal{L}(\textbf{y}, \hat{\textbf{y}}) - \lambda \cdot (d(\textbf{X}_A, f(\textbf{h}_1, \bm{\theta}_{\mathcal{A}d})) + d(\textbf{X}_B, f(\textbf{h}_1, \bm{\theta}_{\mathcal{B}d}))),
\end{equation}
where $\lambda$ is the defender weight, and $\mathcal{L}(\textbf{y}, \hat{\textbf{y}})$ is designed based on different prediction tasks, e.g., $\mathcal{L}$ be the logistic loss for a binary classification task.

\subsection{Learning Model Parameters}
The loss function in Eq. \eqref{obj} is difficult to be solved due to the complex architectures. 
We learn the loss function using iterative optimization method via gradient descent using back propagation \cite{lecun2015deep}, as summarized in Algorithm 2. 
Both forward computation and backward computation need communication between $\mathcal{A}$, $\mathcal{B}$, and the server, in a decentralized manner. 
During training, all the private data ($\textbf{X}_A$, $\textbf{X}_B$, and $\textbf{y}$) and private data related model parameters ($\bm{\theta}_A$, $\bm{\theta}_B$, and $\bm{\theta}_y$) are kept by data holders. Therefore, data privacy is kept to a large extent. 

It is worth noticed that our proposal can be generalized to the situations that the data holders collaboratively calculate $i$ ($1 \le i \le L$) hidden layers instead of the first hidden layer only. Therefore, the existing method \cite{mohassel2017secureml} is one of our special cases, i.e., $\mathcal{A}$ and $\mathcal{B}$ collaboratively calculate all the neural networks using secure multi-party computation techniques, without the neutral server.

\subsection{Implementation}
Communication and computation are two key parts of the decentralized implementation. 
We will describe our solution in details. 

\nosection{Communication}
 The communication includes two parts, the communication between data holder $\mathcal{A}$ and data holder $\mathcal{B}$, and the communication between both data holders and server. 
We adopt Google's gRPC protocol\footnote{https://grpc.io/} to make connection and exchange data between server and data holders. 
Before training, we configure detailed parameters for server and data holders, such as IP addresses, gateways, and dataset locations. 
At the beginning of the training, server and data holders shake hands to build connection. 
After that, they exchange data to finish model training following Algorithm 2.

\nosection{Computation}
The computation are mainly in two parts, i.e., the computations by data holders and the computations by server. 
First, we implement the computations by data holders using Python by ourselves. 
Second, for the heavy computations by server, we choose TensorFlow\footnote{https://www.tensorflow.org/} as backend to perform forward and backward computations. Note that our proposal can be easily implemented by using other deep learning platforms such as PyTorch. 

%% file: section/experiment.tex
\section{Empirical Study}	

\subsection{Experimental Settings}

\nosection{Datasets} 
To test the effectiveness of our proposed model, we choose two public benchmark datasets from Kaggle, both of which are binary classification tasks. The first one is a fraud detection dataset \cite{dal2014learned}, where there are 28 features and 284,807 transactions. The other one is financial distress dataset, where there are 85 features and 3,672 transactions. After we encode the categorical features, there are 556 features in total. 
We assume these features are hold by two parties, and each of them has equal partial features. Moreover, we randomly split the fraud detection dataset into two parts: 80\% as training dataset and the rest as test dataset. 
We also randomly split the financial distress dataset into 70\% and 30\%, since the test dataset needs more samples. 
We repeat experiments five times and report their average results. 

\nosection{Metrics} 
We adopt Area Under the receiver operating characteristic curve (AUC) as the evaluation metric, since both datasets are binary classification tasks. 
In practice, AUC is equivalent to the probability that the classifier will rank a randomly chosen positive instance higher than a randomly chosen negative instance, and therefore, the higher the better. 

\nosection{Hyper-parameters} 
For the Fraud detection dataset, we use a multi-layer perception with 2 hidden layers whose dimensions are [8,8]. We choose Sigmoid as the activation function \cite{jun1995sigmod} and use gradient descent as the optimizer. We set the learning rate to 0.001.
For the Financial distress dataset, we use a multi-layer perception with 3 hidden layers with dimensions [400, 16, 8], we choose Relu as the activation function \cite{HRrelu2000} in the last layer and Sigmoid function in the other layers, use gradient descent as the optimizer, and set the learning rate to 0.006. 

\subsection{Comparison Results}
\begin{figure}
\centering
\subfigure [\emph{Training loss}]{ \includegraphics[width=4cm]{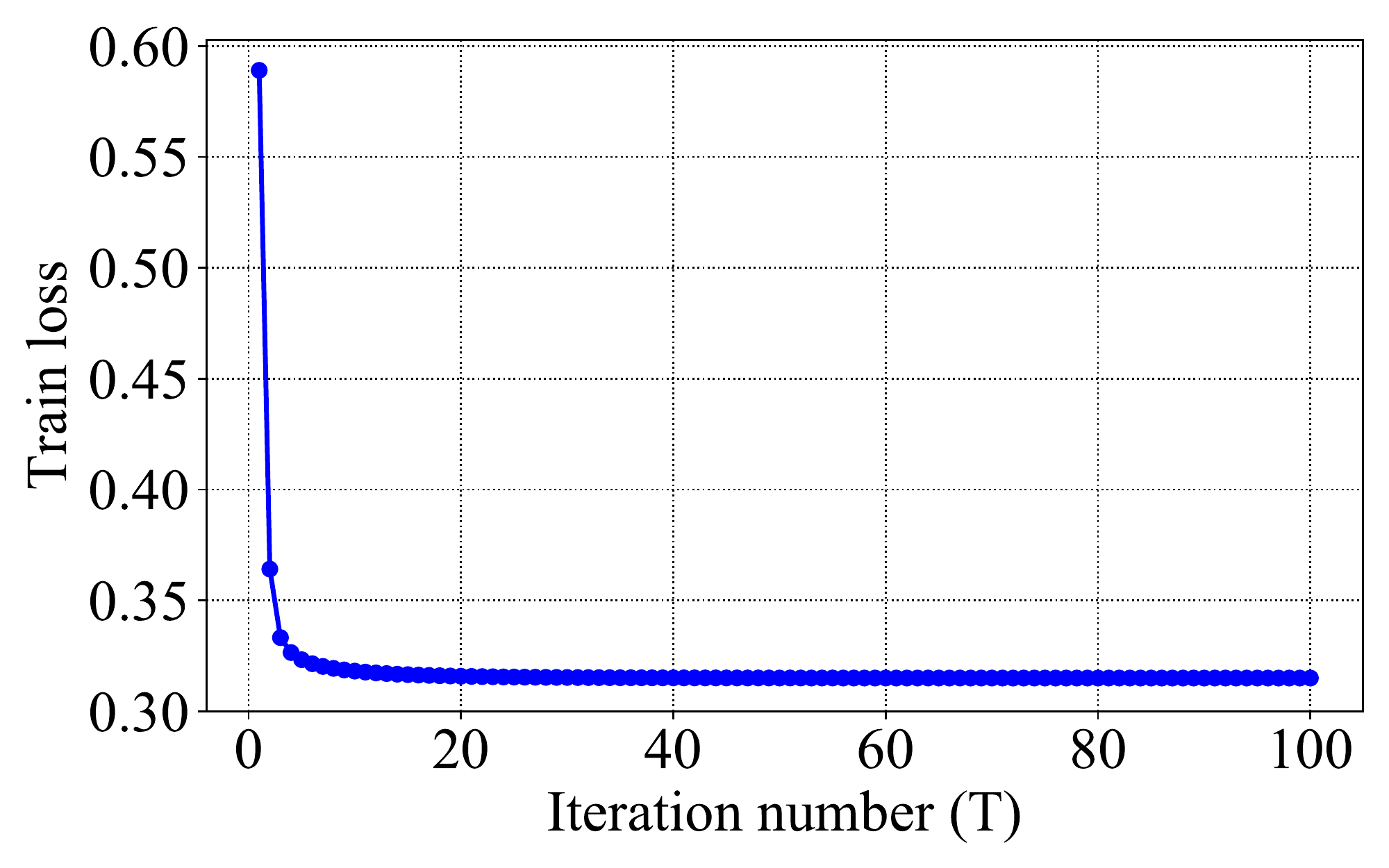}}~~~
\subfigure[\emph{Test loss}] { \includegraphics[width=4cm]{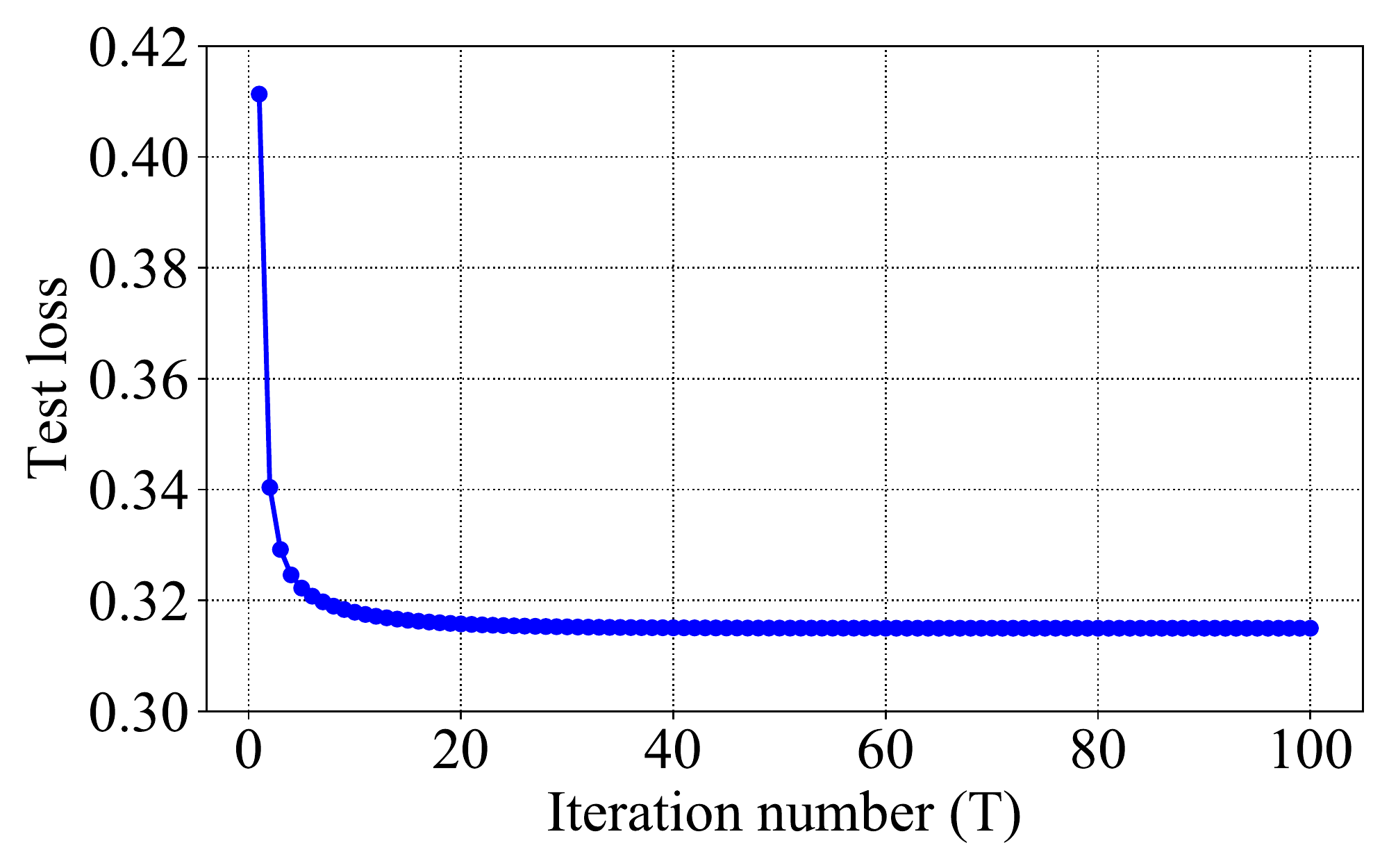}}
\caption{Average loss of $\text{P}^2\text{N}^2$ on fraud detection dataset.}
\label{loss1}
\end{figure}

\begin{figure}
\centering
\subfigure [\emph{Training loss}]{ \includegraphics[width=3.9cm]{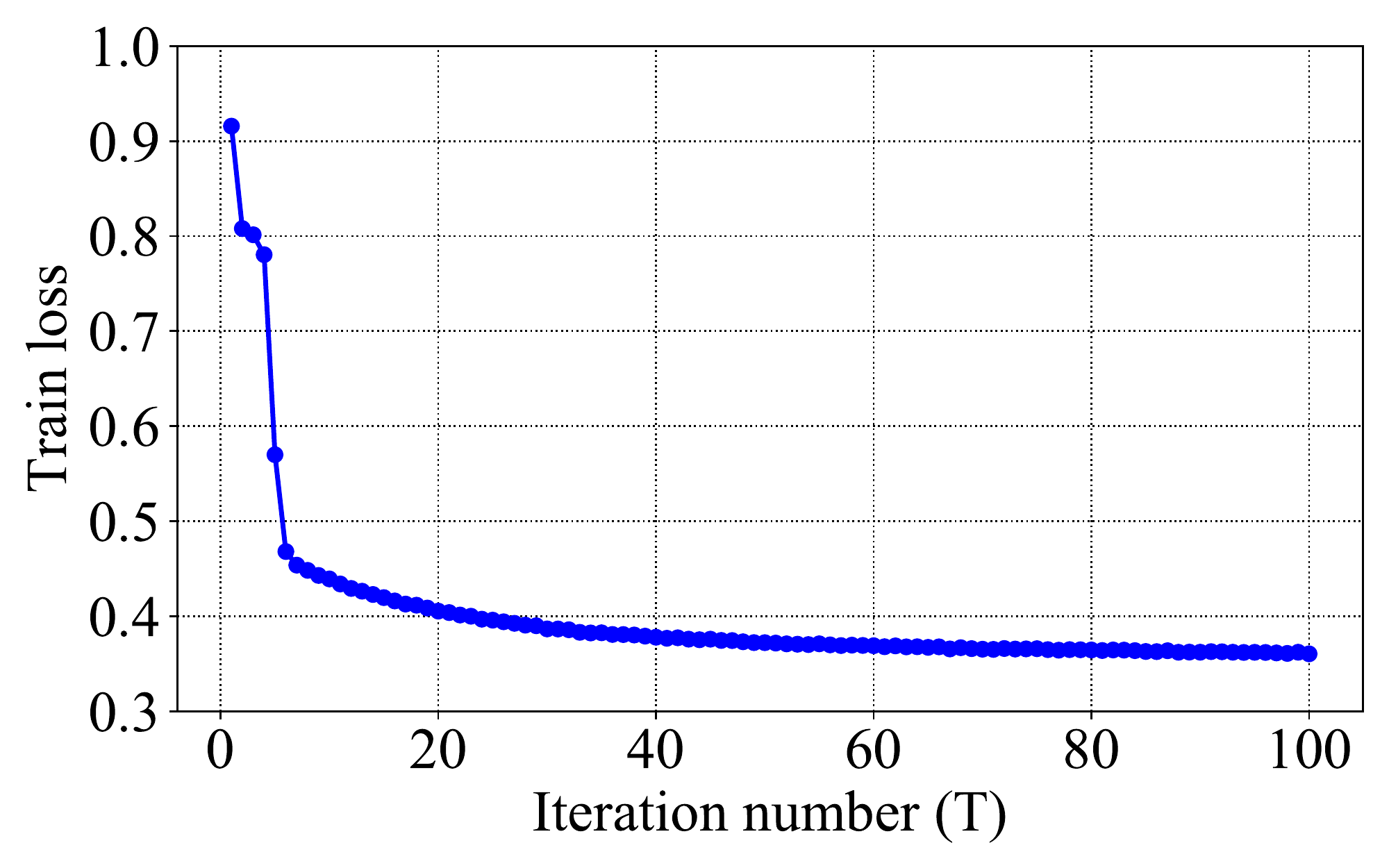}}~~~
\subfigure[\emph{Test loss}] { \includegraphics[width=3.9cm]{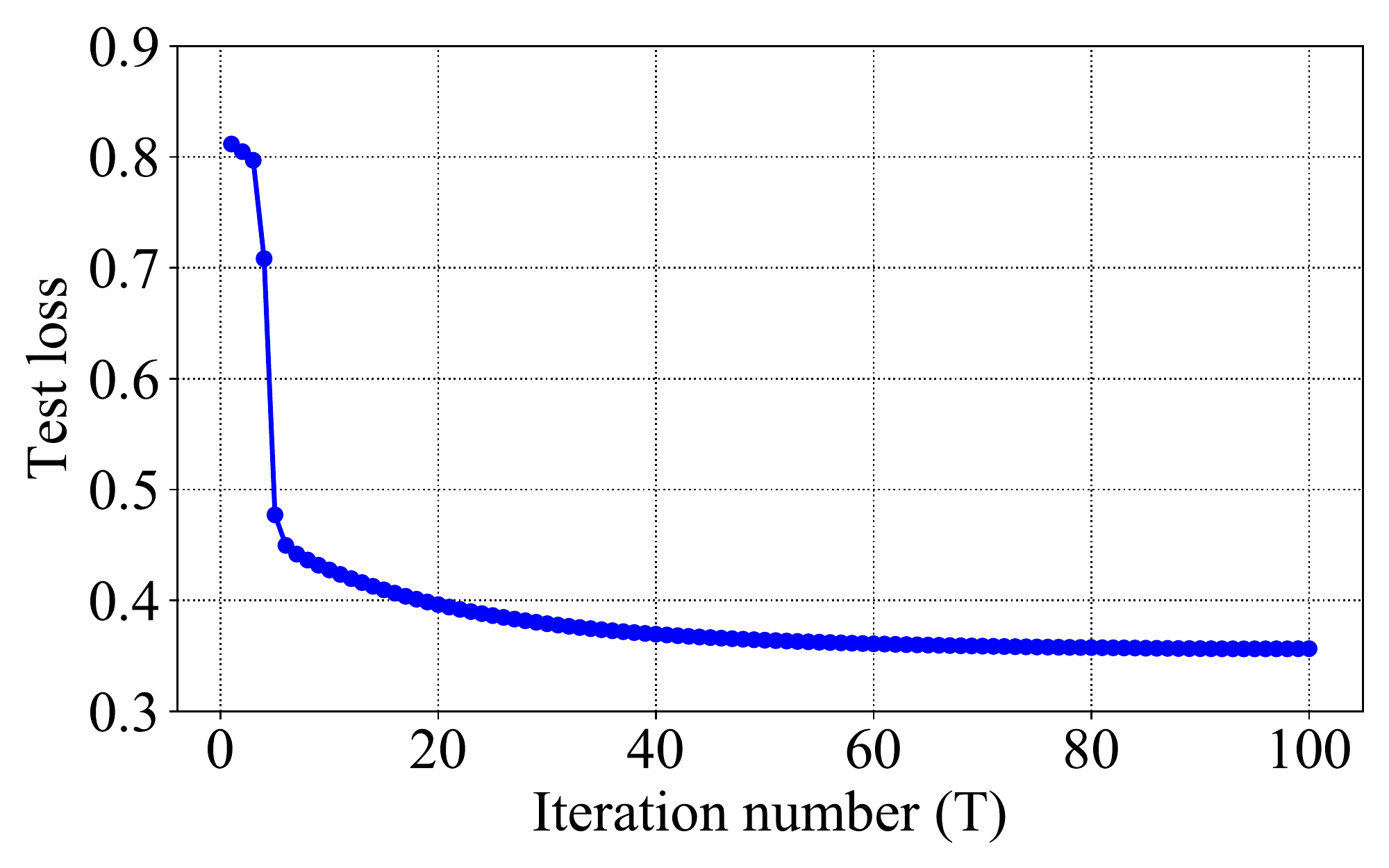}}
\caption{Average loss of $\text{P}^2\text{N}^2$ on financial distress dataset.}
\label{loss2}
\end{figure}

To study the effectiveness of our proposed Privacy Preserving Neural Network ($\text{P}^2\text{N}^2$), we compare it with the traditional neural network (NN) and report the AUC performances on both datasets in Table \ref{compare AUC}, where we set the defender weight $\lambda=0$. 
From it, we can see that $\text{P}^2\text{N}^2$ achieves almost the same prediction performance as NN, which is consistent with the exiting research \cite{mohassel2017secureml}. 
Besides, we show the average training loss and average test loss w.r.t the iteration on two datasets in Figure \ref{loss1} and Figure \ref{loss2}, respectively, where we can see that $\text{P}^2\text{N}^2$ converges steadily without over-fitting. 
The results demonstrate the practicalness of our proposed model. 

\begin{table}
\centering
\caption{Comparison results on two datasets in terms of AUC}
\label{compare AUC}
\begin{tabular}{|c|c|c|}
  \hline
  AUC & NN & $\text{P}^2\text{N}^2$  \\
  \hline
  Fraud Detection & 0.9270 & 0.9231 \\
  \hline
  Financial Distress & 0.9379 & 0.9314 \\
  \hline
\end{tabular}
\end{table}

\subsection{Efficiency Results}

We now study the efficiency of our proposed $\text{P}^2\text{N}^2$, including the comparison of $\text{P}^2\text{N}^2$ and NN, the running time of $\text{P}^2\text{N}^2$ with different training data size and bandwidth.

\nosection{Comparison of training time} 
First, to study the efficiency of ($\text{P}^2\text{N}^2$), we compare the training time of $\text{P}^2\text{N}^2$ and NN on both datasets.
Note that $\text{P}^2\text{N}^2$ is implemented on three PCs which are used as the server and the data-holders in Local Area Network (LAN), and we currently ignore the communication delay between data holders and server.
The results are summarized in Table \ref{compare time}, where we set batch size to 5000.
From it, we find that $\text{P}^2\text{N}^2$ is 70$\times$ slower than NN on the fraud detection dataset and 240$\times$ slower on the financial distress dataset. 
This is because the first hidden layer is calculated by using secret sharing technique instead of plaintext computations, which takes extra communication time, and the first layer dimension on the fraud detection dataset is smaller than that on the financial distress dataset. 
The results indicate that if all the hidden layers are computed using secure multi-party computation techniques, similar as the existing privacy preserving neural networks, the running time will be much longer than our proposal (depend on the depth of the hidden layers). 

\begin{table}[t]
\centering
\caption{Comparison of training time (in seconds) on both datasets}
\label{compare time}
\begin{tabular}{|c|c|c|}
  \hline
  Training time & NN & $\text{P}^2\text{N}^2$  \\
  \hline
  Fraud detection & 21.52 & 1478.32 \\
  \hline
  Financial distress & 5.07 & 1196.67 \\
  \hline
\end{tabular}
\end{table}

\begin{figure}[t]
\centering
\includegraphics[width=6.5cm]{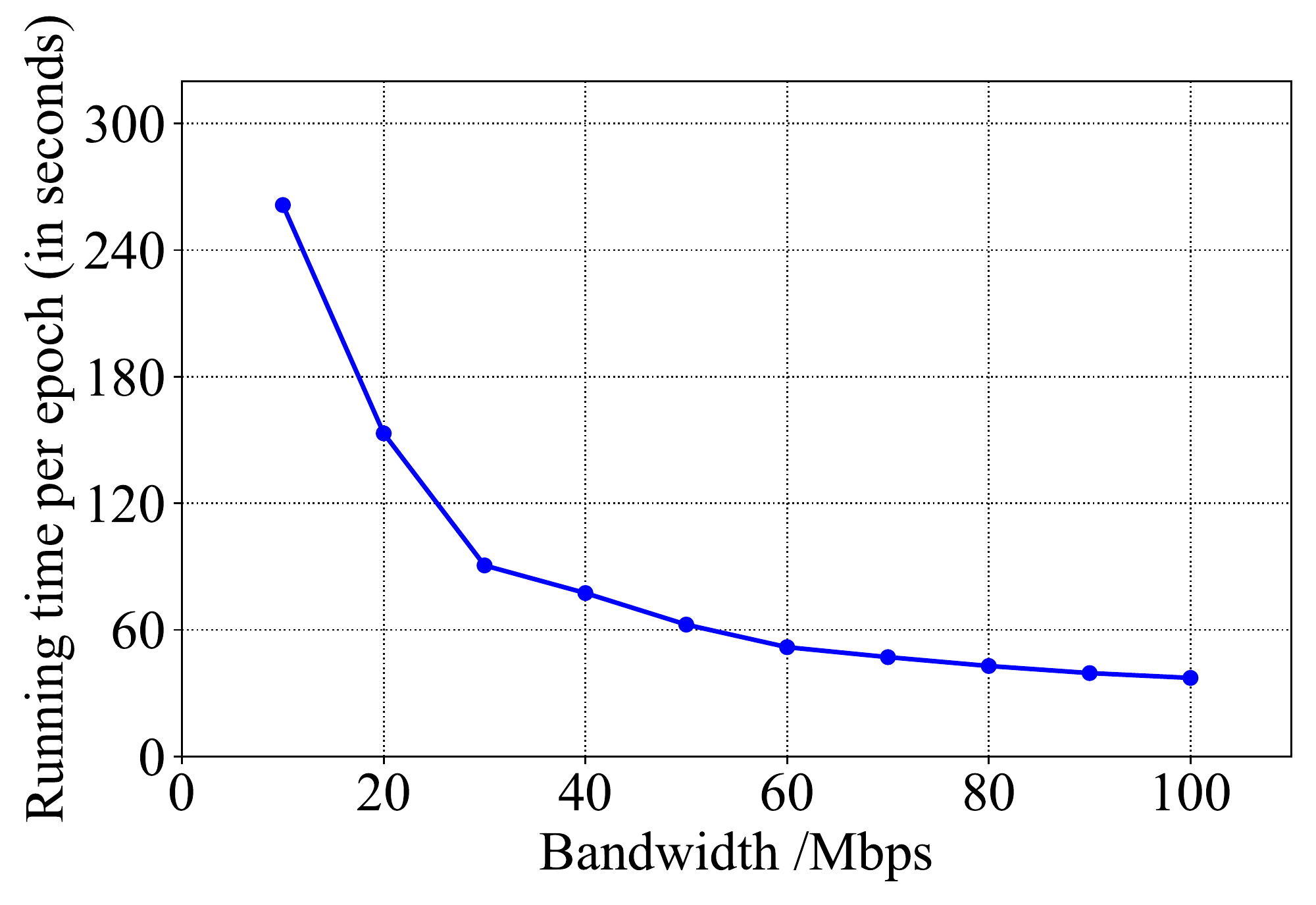}
\caption{Running time of $\text{P}^2\text{N}^2$ with different bandwidth.}
\label{mpc_time_bandwidth}
\end{figure}

\nosection{Running time with different bandwidth} 
Second, we study the training time of $\text{P}^2\text{N}^2$ on the fraud detection dataset by varying network bandwidth. The result is shown in Figure \ref{mpc_time_bandwidth}. 
From it, we find that with the increase of network bandwidth, the training time of $\text{P}^2\text{N}^2$ first rapidly decreases and then tends to be stable. 
The result indicates that the efficiency of our proposed $\text{P}^2\text{N}^2$ heavily relies on the network status. 

\begin{figure}[t]
\centering
\includegraphics[width=6.5cm]{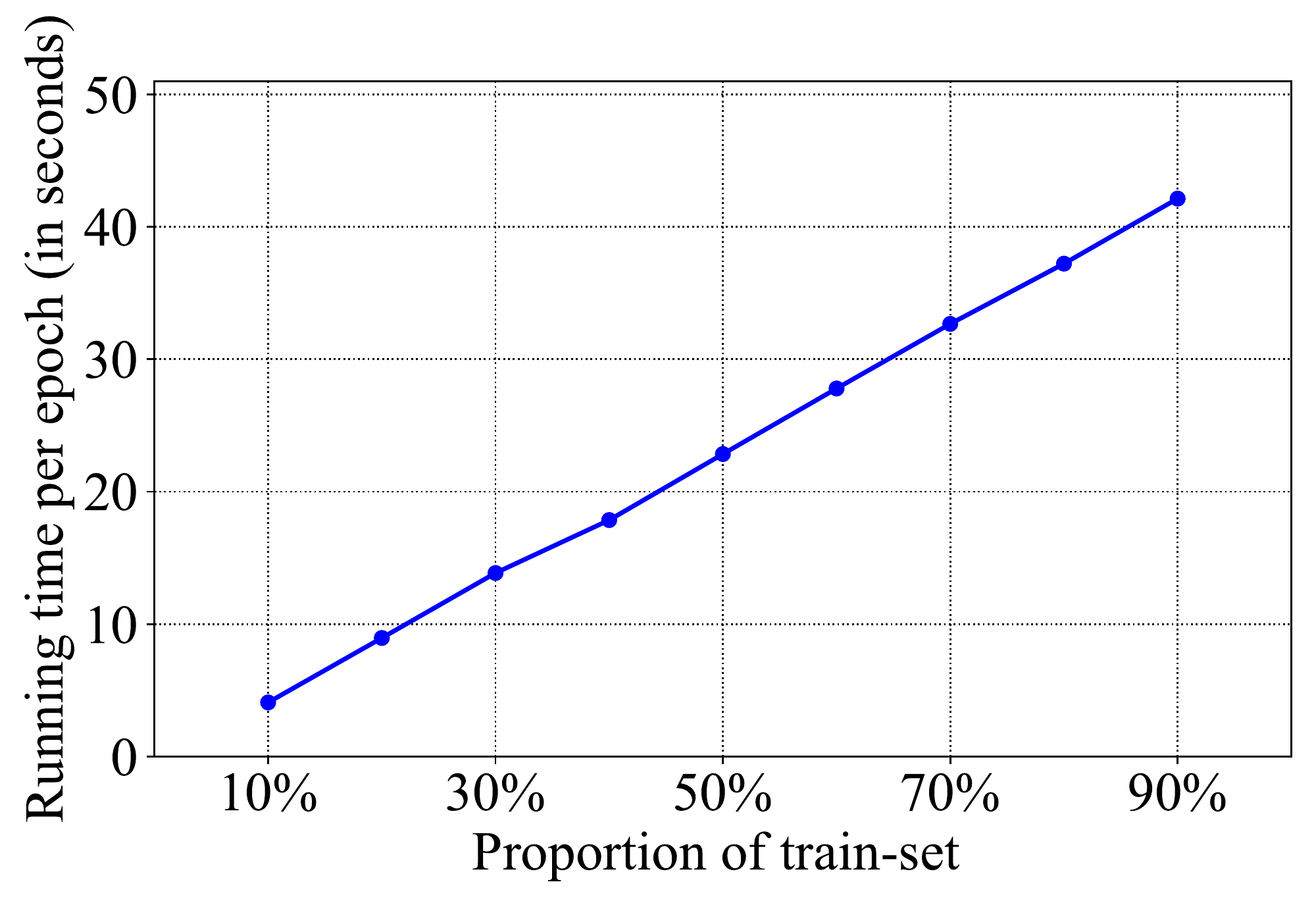}
\caption{Running time of $\text{P}^2\text{N}^2$ with different datasize.}
\label{mpc_time_data}
\end{figure}

\nosection{Running time with different data size} 
Furthermore, we study the running time of $\text{P}^2\text{N}^2$ with different data size, where fix the network bandwidth to 100 M/bps.
We do this by varying the proportion of training data size using the fraud detection dataset, and report the running time of $\text{P}^2\text{N}^2$ in Figure \ref{mpc_time_data}. 
From it, we find that the running time of $\text{P}^2\text{N}^2$ scales linearly with the training data size. 
The results indicate that our proposed $\text{P}^2\text{N}^2$ can be scale to large dataset. 

\subsection{Effect of Defender}
We finally study the effects of the defender on both model accuracy and its privacy preserving ability. 

\begin{figure}[t]
\centering
\includegraphics[width=6.5cm]{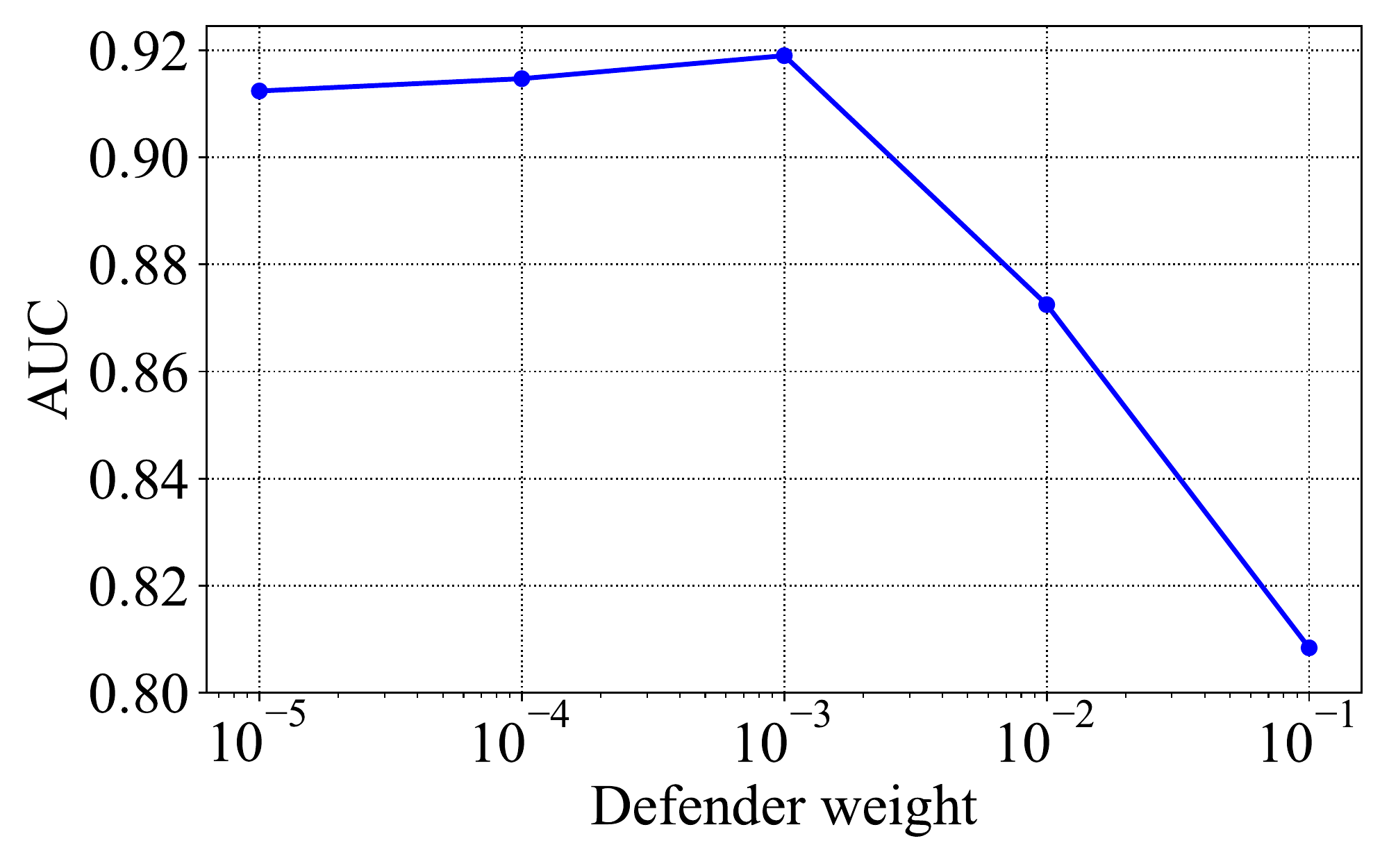}
\caption{Effect of defender weight $\lambda$ on $\text{P}^2\text{N}^2$.}
\label{effect-lambda}
\end{figure}

\nosection{Influence on accuracy}
We first vary the defender weight $\lambda$ in $\{ 10^{-5}, 10^{-4},...,10^{-1} \}$ and study its influence on our model performance, where we use the fraud detection dataset. 
We report the results in Figure \ref{effect-lambda}. 
Note that $\lambda=0$ indicates the absence of such a defender. 
We observe that, with the increase of $\lambda$, the accuracy of $\text{P}^2\text{N}^2$ first slightly increases and then quickly decreases. 
This is because, the objective function in Eq. \eqref{obj} has two parts, i.e., the cross-entropy loss that determines model accuracy and the defender loss which determines the privacy preserving ability, and they are balanced by $\lambda$. 
When $\lambda$ is a small value (but bigger than 0), it works like a penalty term which prevents the model from overfitting to a certain extent. However, when $\lambda$ is too big, $\text{P}^2\text{N}^2$ pays more attention to the defender loss rather than the cross-entropy loss. Therefore, the accuracy starts to decrease quickly. 

\nosection{Influence on privacy preserving ability}
Second, to visually demonstrate the effectiveness of our proposed defender strategy, we conduct the following experiments on the MNIST dataset---a handwritten digits classification dataset \cite{lecun1998mnist}, where each private input record is a handwritten digit from 0 to 9. 
During private input recovery experiments, we assume a serious private information leakage situation, i.e., the neutral server obtains some of the private input and the corresponding hidden layer of the training dataset. 
Based on these leaked information, the server can learn an attacker that maps the hidden layer to private input. 
After it, the attacker can easily recover the input of other records given their hidden layers. 

We compare the recovery result with and without the defender. We use a three-layer fully-connected neural network, i.e., a multi-layer perception, as the defender. The network structure of the defender is (728, 512, 128, 10), where 728 is the dimension of each handwritten digit, 512 and 128 are hidden layer dimensions, and 10 is the output dimension (classification number). 
During experiments, we assume two data holders have evenly partial features of a digit. 
We choose Relu \cite{lecun2015deep} as the active function, mean squared error as the distant function $d(\cdot,\cdot)$ in Eq. \eqref{obj}, Adam \cite{kingma2014adam} as the optimizer, and set learning rate to 0.01. 

\begin{figure*}
\centering
\subfigure [Original input]{ \includegraphics[width=5cm]{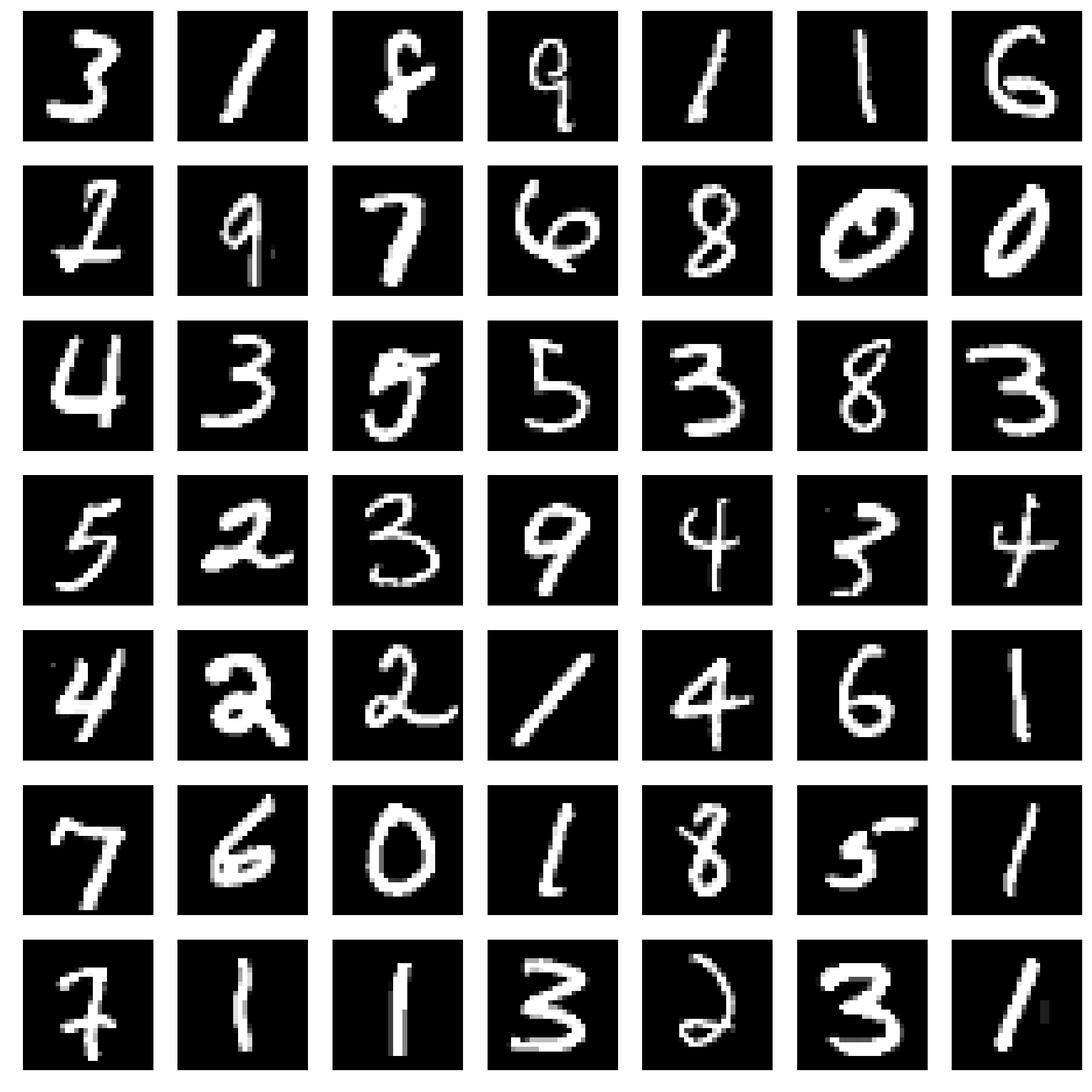}}~~~
\subfigure [Without defender]{ \includegraphics[width=5cm]{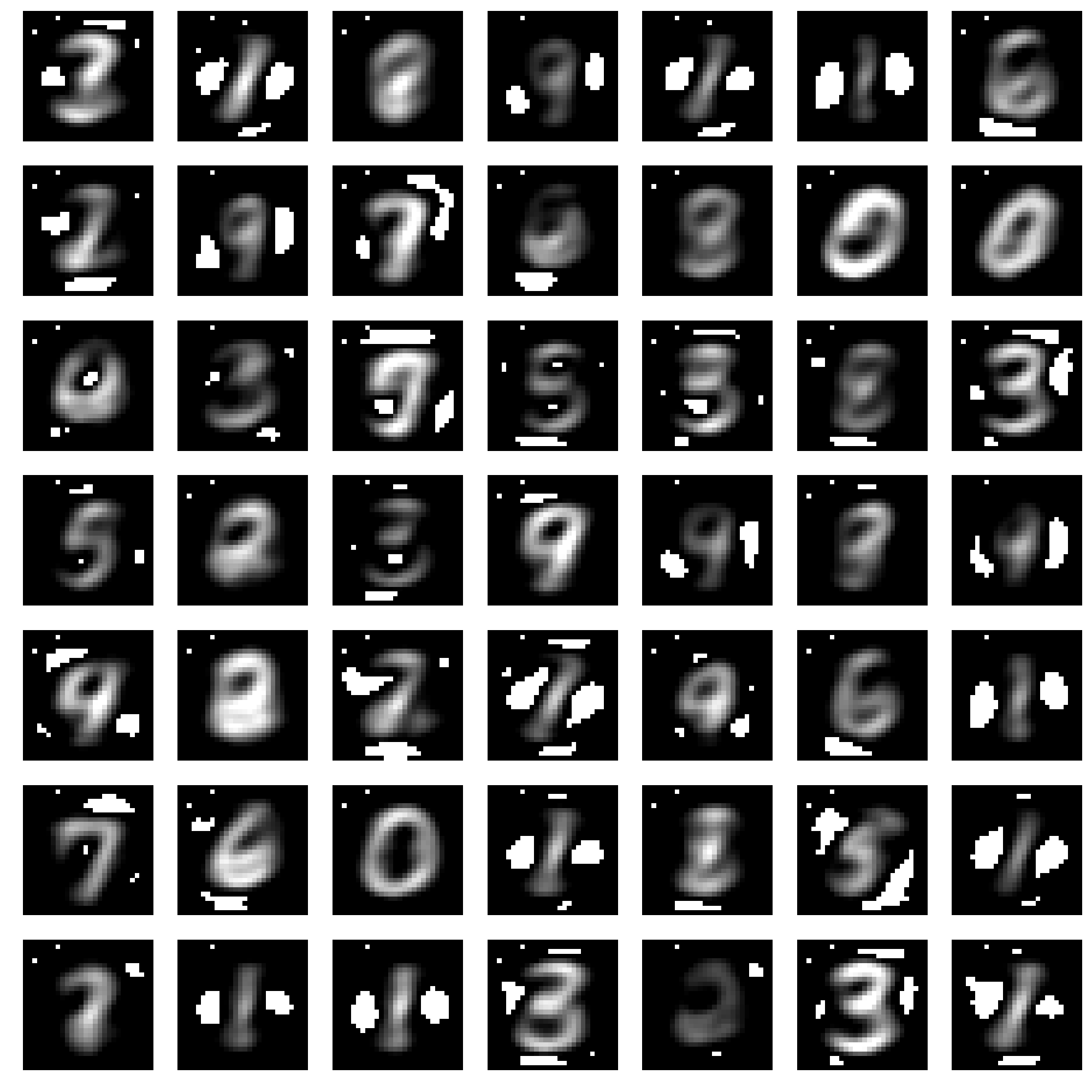}}~~~
\subfigure[With defender] { \includegraphics[width=5cm]{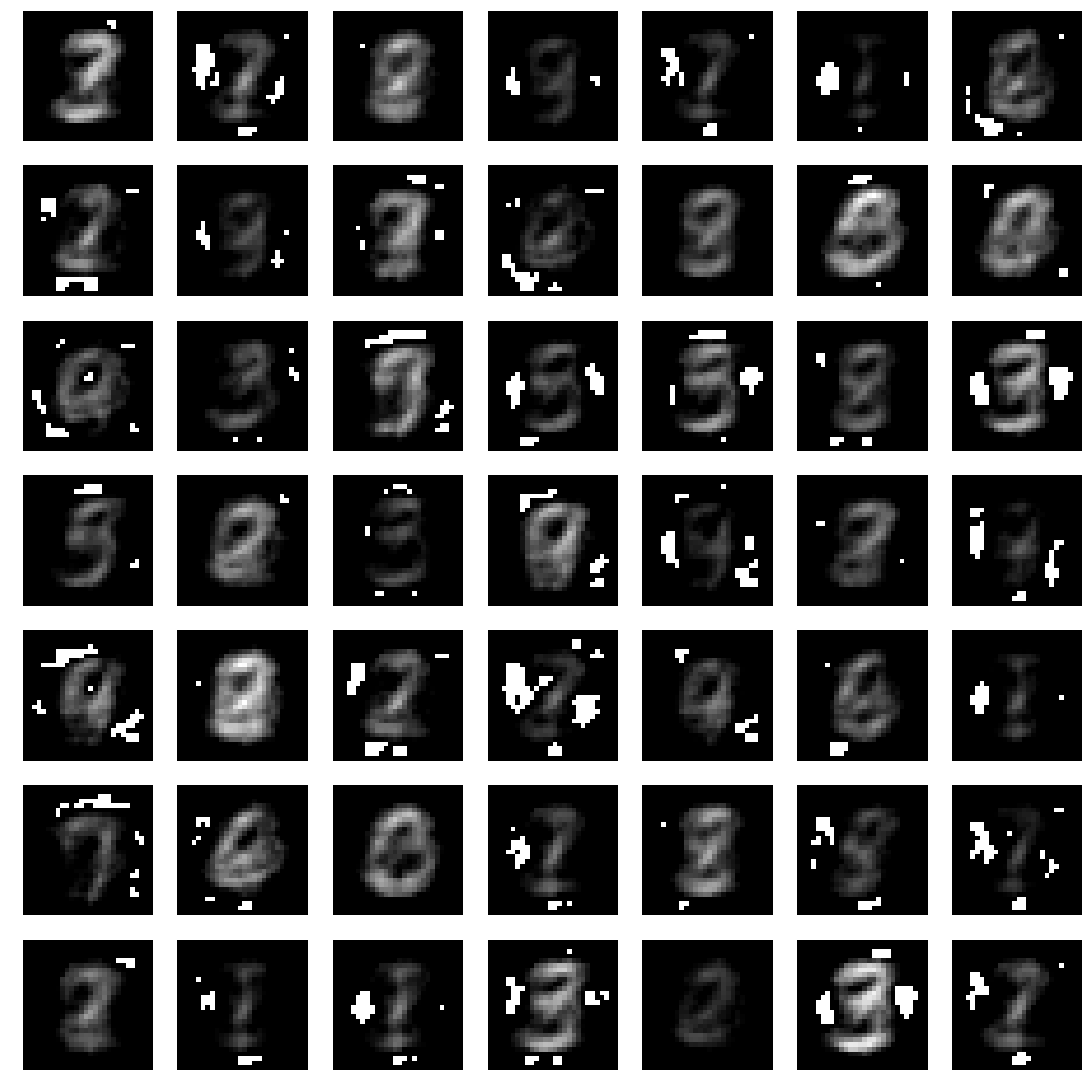}}
\caption{Private input recovery results on MNIST dataset. (a) is the randomly selected original handwritten digits, (b) is the corresponding recovered result by the server without the presence of the defender, and (c) is the recovered result by the server but with the presence of the defender on data holders. }
\label{defender}
\end{figure*}

We report the input recovery result in Figure \ref{defender}, where (a) is the randomly selected original handwritten digits, (b) and (c) are the corresponding recovered results with and without the defender, respectively. 
We set the defender weight $\lambda=100$ on MNIST dataset, since the attack loss is small comparing with that on fraud detection dataset. 
From it, we can clearly see that, with the presence of the defender, it becomes more difficult to recognize the recovered digits. 
The results indicate that the defender mechanism can effectively preventing the server from recovering the private input of data holders. 

Note that one can still make a good guess on the recovered digits from Figure \ref{defender} (C), even with the defender mechanism. 
This is because, in our experiments, we assumed that the neutral server has obtained some private input from data holders. 
This is quite serious data leakage situation and is very difficult to appear in practice, since these private input are hold by different data holders. 
Nevertheless, our experiments demonstrated the effectiveness of the defender mechanism. 

%% file: section/conclusion.tex
\section{Conclusion}
In this paper, we proposed a privacy preserving neural network learning paradigm that can scale to large datasets. 
Our motivation is to split the computation graph of DNN into two parts, i.e., the computations related to private data are performed by data holders using cryptographical techniques, and the rest of the computations are done by a neutral server with high computation ability. 
Our model achieved promising results on real-world fraud detection dataset and financial distress dataset. 
In the future, we would like to deploy our proposal in real-world applications in Ant Financial.